\documentclass[12pt,a4paper]{article}

\usepackage[utf8x]{inputenc}
\usepackage[T1]{fontenc}

\usepackage[a4paper,top=2cm,bottom=2cm,left=3cm,right=3cm,marginparwidth=1.75cm]{geometry}

\usepackage[affil-it]{authblk} 

\usepackage{amsmath}
\usepackage{graphicx}
\usepackage[colorinlistoftodos]{todonotes}
\usepackage{mathptmx}
\usepackage{titlesec}
\usepackage{multirow}
\usepackage{booktabs}
\usepackage[colorlinks=true, allcolors=blue]{hyperref}
\usepackage[nameinlink]{cleveref}

\titleformat{\section}
  {\normalfont\fontsize{12}{15}\bfseries}{\thesection.}{1em}{}
\titleformat{\subsection}
  {\normalfont\fontsize{12}{15}\bfseries\itshape}{}{0.1em}{}

\titlespacing*{\section}
  {0pt}{1\baselineskip}{0\baselineskip}
\titlespacing*{\subsection}
  {0pt}{0\baselineskip}{0\baselineskip}

\usepackage{caption}

\usepackage{apacite}
\usepackage[english]{babel}

\addto{\captionsenglish}{%
    }
\title{\bfseries Pedestrian 3D Bounding Box Prediction}

\author[1]{Saeed Saadatnejad}
\author[1]{Yi Zhou Ju}
\author[1]{Alexandre Alahi}

\affil[1]{Visual Intelligence for Transportation lab (VITA), Ecole Polytechnique Federale de Lausanne (EPFL), Switzerland}

\date{\vspace{-5ex}}

\begin{document}
\maketitle
\section*{SHORT SUMMARY}
\small

Safety is still the main issue of autonomous driving, and in order to be globally deployed, they need to predict pedestrians' motions sufficiently in advance.
While there is a lot of research on coarse-grained (human center prediction) and fine-grained predictions (human body keypoints prediction), we focus on 3D bounding boxes, which are reasonable estimates of humans without modeling complex motion details for autonomous vehicles. This gives the flexibility to predict in longer horizons in real-world settings.
We suggest this new problem and present a simple yet effective model for pedestrians' 3D bounding box prediction.
This method follows an encoder-decoder architecture based on recurrent neural networks, and our experiments show its effectiveness in both the synthetic (JTA) and real-world (NuScenes) datasets. The learned representation has useful information to enhance the performance of other tasks, such as action anticipation.
Our code is available online. \footnote{\url{https://github.com/vita-epfl/bounding-box-prediction}}

\textbf{Keywords}: 3D Bounding Box Prediction, Autonomous Vehicles, Motion Prediction

\section{Introduction}
Predicting the future location of a pedestrian is key to safe-decision making for autonomous vehicles (AVs). It is a non-trivial task for AVs because humans can choose complex paths and move at non-uniform speeds. Furthermore, AVs should predict a pedestrian's location sufficiently in advance to react accordingly. Consider a scenario where a self-driving car identifies a person standing on the sidewalk near an intersection. If the vehicle can predict the pedestrian's future locations, it can better decide when to stop to ensure pedestrian safety and not be too conservative. In the application of AVs, there are different levels of pedestrian motion prediction (coarse or fine-grained).

Predicting a sequence of future center positions (x-y coordinates) given a set of observed positions were widely studied in the literature by model-based approaches \cite{helbing1995social,pellegrini2009you} and data-driven approaches \cite{kothari2020human, saadatnejad2021sattack, bahari2021vehicle, mohamed2020social} to name a few.
However, the coarse-grained information is insufficient for an AV to make a safe planning decision without unnecessary cautions.
Some works predict fine-grained human motions such as body keypoints. Indeed, instead of predicting $1$ point per pedestrian, they predict $17$ or more keypoints \cite{parsaeifard2021decoupled, long_term, synthesizing_long_term}. 
Considering this high-dimensional complex data, it is hard for the predictor to accurately forecast for long periods in real-world settings. Note that in most real-world applications, human keypoints would not be readily available but would need to be estimated by another model first, which could be noisy.
Another line of research predicts future center locations $(x,y)$ along with the width and height $(w,h)$ of humans given the past information. We refer to this task as the bounding box prediction \cite{bhattacharyya2018long, styles2020mof, bouhsain2020pedestrian}. 
It provides a strong high-level intermediate representation of human motions without modeling unnecessary details for AVs. Thus, it can be used for longer prediction horizons in the real world.
Note that it is easier and more accurate to estimate bounding boxes \cite{park2021detection3d}.

Most of the research in predicting pedestrian motions is in two dimensions (2D), which is likely due to the abundance of datasets that provide comprehensive 2D annotations of pedestrians in the wild. For instance, the 2D Joint Attention in Autonomous Driving (JAAD) dataset provides bounding box annotations from the vehicle's egocentric view \cite{rasouli2017they}.
3D information can capture more complexities of human motion and provide better estimates of pedestrian location and size for AVs, however, it requires more costs compared to 2D.
Recently, with the advancements in devices and data collection, few datasets with full 3D annotations of pedestrians in real-world scenarios became accessible \cite{nuscenes, kim2019pedx}.
In addition, we can benefit from some simulated synthetic 3D  datasets, as they provide 100\% accurate annotations. They use simulated environments to provide 3D annotations automatically \cite{JTA}.

This project tackles the problem of predicting pedestrian 3D bounding boxes $(x,y,z,w,h,d)$.
We propose a simple yet effective encoder-decoder Long-Short Term Memory (LSTM) network that has two encoders and two decoders similar to \cite{bouhsain2020pedestrian}, both of which are components proven to be effective for multiple computer vision tasks.
The encoder comprises two parallel LSTMs to process 3D bounding box locations and velocities, and the decoder has two LSTM, one for future 3D bounding box prediction and the other for attribute detection.
We demonstrate the effectiveness of the model in both synthetic and real-world datasets and the usefulness of the learned representation in action anticipation.
With this research, we encourage the community to conduct more research in human 3D bounding box prediction.

\section{Method} \label{sec:methods}

\begin{figure}[!t]
    \centering
    \includegraphics[width=\textwidth]{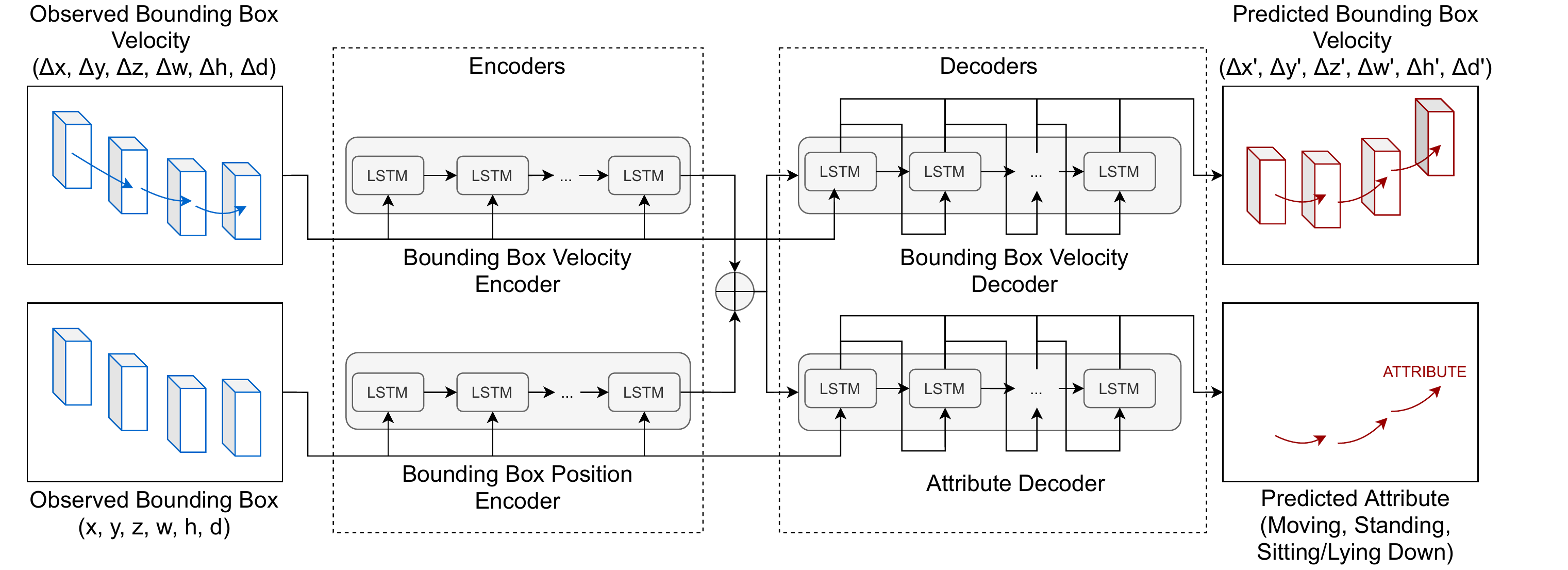}
    \caption{The proposed encoder-decoder architecture}
    \label{fig:network architecture}
\end{figure}

The proposed network is a sequence to sequence LSTM model called the Position-Velocity LSTM (PV-LSTM), as it encodes both the position and the velocity of the 3D pedestrian bounding box, see \Cref{fig:network architecture}. The input to the network is the observed speed and position of the 3D bounding box center along with its weight, height, and depth. The output is the predicted speed of the 3D bounding box elements, which can then be converted into a corresponding absolute position. The network consists of LSTM encoder-decoders for position and speed, followed by a fully connected layer. Given a sequence of bounding boxes of a single pedestrian $(p_{t-T_{obs}+1},...,p_t)$ corresponding to the time-steps or frames $(t-T_{obs}+1,...,t)$ in a video sequence, the network outputs the next sequence of bounding boxes $(p_{t+1},...,p_{t+T_{pred}})$  at the following time-steps $(t+1,...,t+T_{pred})$. The 3D bounding box that encompasses a pedestrian at time step $t$ is represented by the coordinates of its center and its width, height, and depth $p_t = (x_t, y_t, z_t, w_t, h_t, d_t)$.

\subsection{Bounding box velocity encoder}
The bounding box velocity encoder $LSTM_{v}^{enc}$ encodes the velocity of the observed bounding boxes. The velocities are represented as the difference between successive pairs of bounding box centers and sizes, where $v_{t}=(\Delta x_t, \Delta y_t, \Delta z_t, \Delta w_t, \Delta h_t, \Delta d_t)=(x_t-x_{t-1}, y_t-y_{t-1}, z_t-z_{t-1}, w_t-w_{t-1}, h_t-h_{t-1}, d_t-d_{t-1})$. Given the observed velocities as input, the LSTM encoder computes the hidden state of the velocity $v_t$ as time $t$ using:
\begin{equation}
    {hv}_t=LSTM_{v}^{enc}({hv}_{t-1},v_t).
\end{equation}

\subsection{Bounding box position encoder}
The bounding box position encoder $LSTM_{p}^{enc}$ encodes the positions and dimensions of the observed bounding boxes. It takes the observed bounding box information $(p_{t-T_{obs}+1},...,p_t)$ and outputs the updated position hidden state $p_t$ at time $t$ using:
\begin{equation}
    {hp}_t=LSTM_{p}^{enc}({hp}_{t-1},p_{t}).
\end{equation}

\subsection{Bounding box velocity decoder}
The bounding box velocity decoder $LSTM_{v}^{dec}$ is used to predict the next sequence of bounding box velocities \(\hat{v}_{t+1}=(\Delta \hat{x}_{t+1}, \Delta \hat{y}_{t+1}, \Delta \hat{z}_{t+1}, \Delta \hat{w}_{t+1}, \Delta \hat{h}_{t+1}, \Delta \hat{d}_{t+1})\).

First, the hidden position and velocity states are grouped into one hidden state $h_t$ which combines all the features:
\begin{equation}
    hh_t = hp_t \oplus hv_t.
\end{equation}
Then, the predicted velocity at time $t+1$ is calculated using:
\begin{equation}
\begin{split}
    \hat{hh}_{t+1} &= LSTM_{v}^{dec}(hh_t,v_t), \\
    \hat{v}_{t+1} &= FC (\hat{hh}_{t+1}),
\end{split}
\end{equation}
where $hh_t$ is the initial hidden state, $v_t$ is the last observed velocity, and $FC$ represents the fully connected layer.
The following velocities are then found iteratively.
Then, it is straightforward to compute the absolute predicted bounding boxes given the initial absolute coordinates of the bounding box at time $t$.

\subsection{Attribute decoder}
The attribute decoder $LSTM_{a}^{dec}$ is used to predict the next sequence of attributes $(\hat{a}_{t+1},...,\hat{a}_{t+T_{pred}})$. The architecture is similar to the previous decoder except that the output is passed through a softmax layer. The aim of this decoder is to have accurate attribute anticipation if labels are provided in the training dataset.

\section{Experiments}

\subsection{Datasets}

\textbf{Joint Track Auto (JTA) \cite{JTA}:}
a synthetic dataset for pedestrian pose estimation and tracking in urban scenarios made from photo-realistic video game Grand Theft Auto V (GTA). It includes $512$ videos, where $256$ were used for training, $128$ for validation, and $128$ for testing. Each video is $30$ seconds long with $30$ frames per second (fps), making it a large dataset. The dataset provides the 3D coordinates of humans' keypoints in meters in the camera coordinate system. The 3D bounding box information was extracted from the 3D joint data for each pedestrian in each frame.

The dataset includes both indoor and outdoor scenes featuring many pedestrians. Most of the video sequences are recorded with a stationary camera from various locations with different viewpoints. There are a few sequences recorded with a rapidly moving camera.
The observation length is $0.5$ seconds, and the prediction horizons are $0.5$ seconds and $2$ seconds.

\textbf{NuScenes \cite{nuscenes}:} a large-scale real-world dataset of $1000$ driving scenes collected in the urban cities of Boston and Singapore, each $20$ seconds long with $2$ fps. $850$ scenes were manually annotated for $23$ object classes, such as pedestrians and vehicles, and includes visibility, activity, and pose attributes. These were divided into a train/validation/test split of 550/150/150 sizes. The other $150$ scenes that are not annotated were not used. 

The bounding box information is provided in the coordinate system of that specific camera. Note that the vehicle is mostly stationary, such as stopped at a traffic light or parked, during the collection of the scenes, but there can still be minor movements from frame to frame within each scene.
The observation length and prediction horizon are both $2$ seconds.

\subsection{Evaluation Metrics}
The Average Displacement Error (ADE) and the Final Displacement Error (FDE) are used to evaluate the performance of the predicted bounding box center predictions compared to the ground truth. In addition, to have a better estimate of 3D prediction performance, we compare the volume of the predicted 3D bounding box and the ground truth and compute the Average and Final Intersection volume Over Union (AIOU, FIOU) between them.

\subsection{Implementation Details}
The network was implemented in PyTorch and trained on two NVIDIA GTX-1080-Ti GPUs. An Adam optimizer was used with a learning rate scheduler with the starting learning rate of $1e-3$, a reduction factor of $0.1$, patience of $10$, and a threshold of $1e-8$. The training is done for $100$ epochs with Mean Squared Error (MSE) loss for the 3D bounding box decoder and cross-entropy loss for the attribute decoder.
Our model uses one-layer LSTMs with hidden layers of size $512$ for both the encoder and the decoder.
Other details can be found in the source code.

\subsection{Baselines}
\begin{enumerate}
    \item PV-LSTM: the proposed method
    \item P-LSTM: excluding the velocity encoder of the proposed method. Indeed, only the velocities of bounding boxes are fed and predicted.
    \item Zero velocity (Zero-Vel): a simple baseline to keep the last observed bounding box as the prediction for future frames. It has been shown that this baseline is a hard-to-beat baseline that has outperformed a large number of models in motion prediction \cite{martinez2017human, fragkiadaki2015recurrent}.
\end{enumerate}

\subsection{Results}

The qualitative results on the JTA dataset on short and long horizons are presented in \Cref{tab:jta}. It shows the superiority of the proposed method (PV-LSTM) over the baselines with AIOU/FIOU of $0.6410/0.4809$ for short and $0.3804/0.1843$ for long prediction horizon.
In addition, it shows that excluding the velocity encoder degrades the performance.

\begin{table}[!h] 
    \centering
    \caption{Quantitative results on JTA dataset. For ADE/FDE, the lower the better and for AIOU/FIOU, the higher the better.}
    \begin{tabular}{c| c c c c | c c c c}
        \toprule
        & \multicolumn{4}{c|}{prediction horizon = 0.5 sec} & \multicolumn{4}{c}{prediction horizon = 2 sec} \\
        \midrule
        Model & ADE & FDE & AIOU & FIOU & ADE & FDE & AIOU & FIOU \\
        \midrule
        Zero-Vel & 0.6168 & 1.1327 & 0.3306 & 0.1792 & 1.8364 & 3.5263 & 0.2268 & 0.1778 \\
        P-LSTM & 0.2436 & 0.4440 & 0.5762 & 0.4063 & 0.6409 & 1.2468 & 0.3281 & 0.1431\\
        PV-LSTM & \textbf{0.2187} & \textbf{0.4015} & \textbf{0.6410} & \textbf{0.4809} & \textbf{0.5942} & \textbf{1.1705} & \textbf{0.3804} & \textbf{0.1843}\\
        \bottomrule
    \end{tabular}
    \label{tab:jta}
\end{table}

\begin{figure}[!h]
\centering
\includegraphics[width=0.24\textwidth]{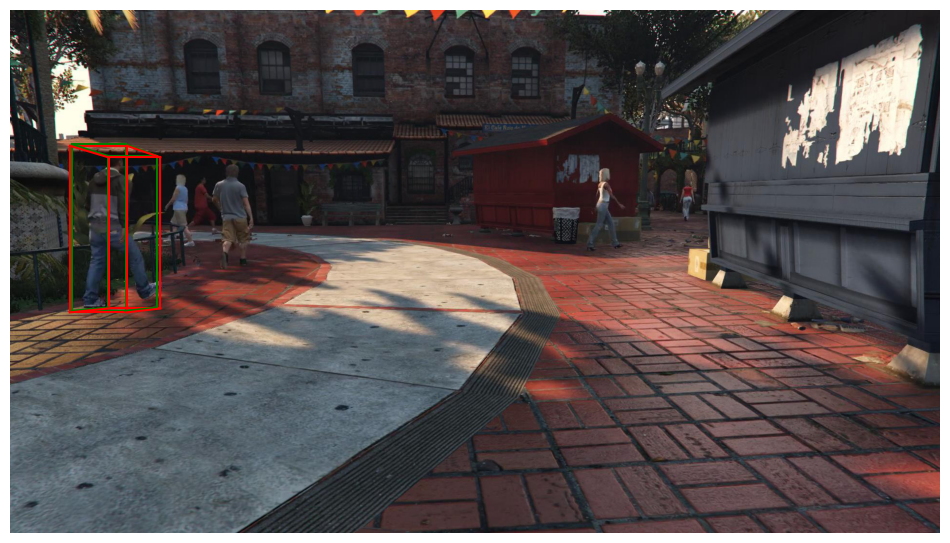}
\includegraphics[width=0.24\textwidth]{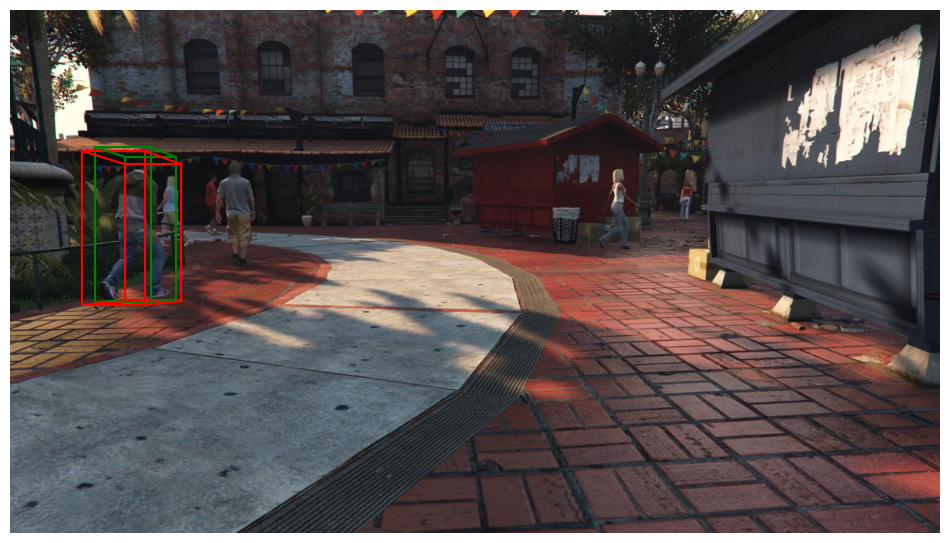}
\includegraphics[width=0.24\textwidth]{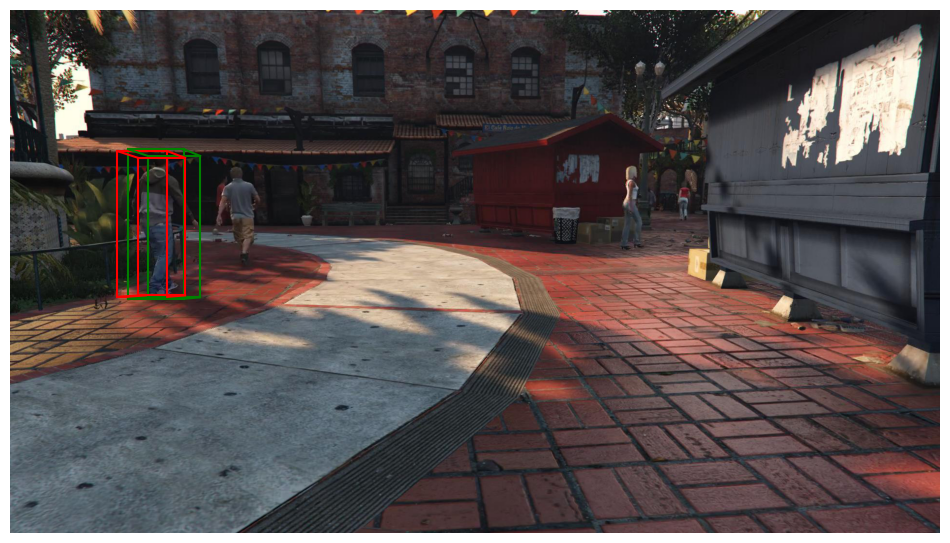}
\includegraphics[width=0.24\textwidth]{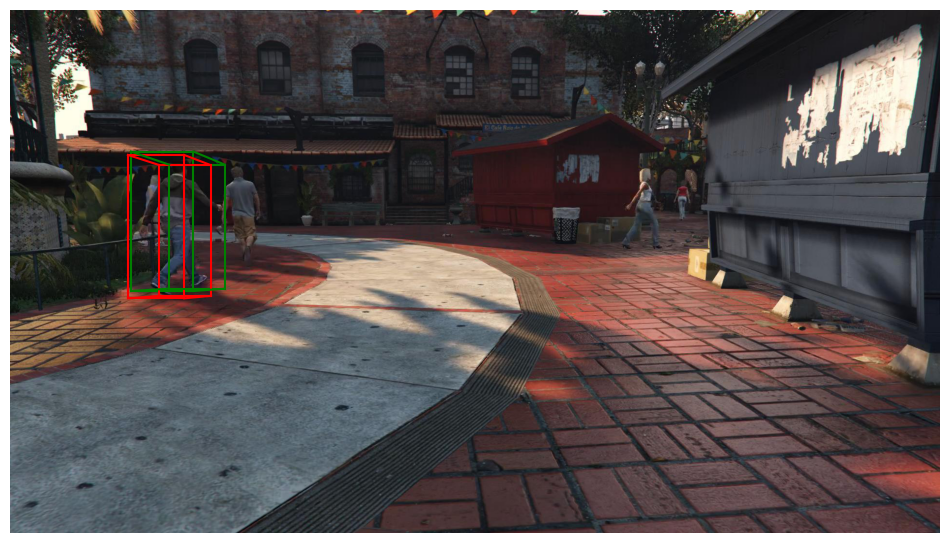}\\
\includegraphics[width=0.24\textwidth]{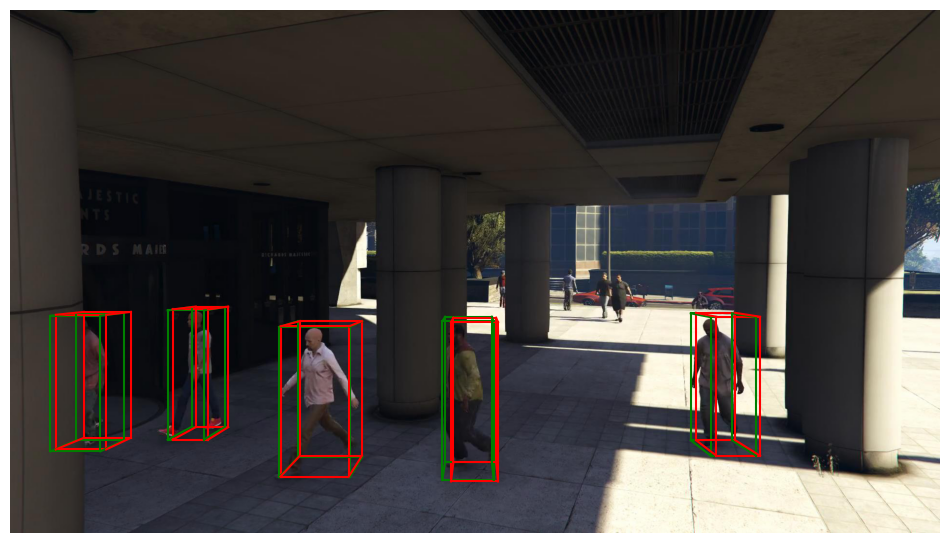}
\includegraphics[width=0.24\textwidth]{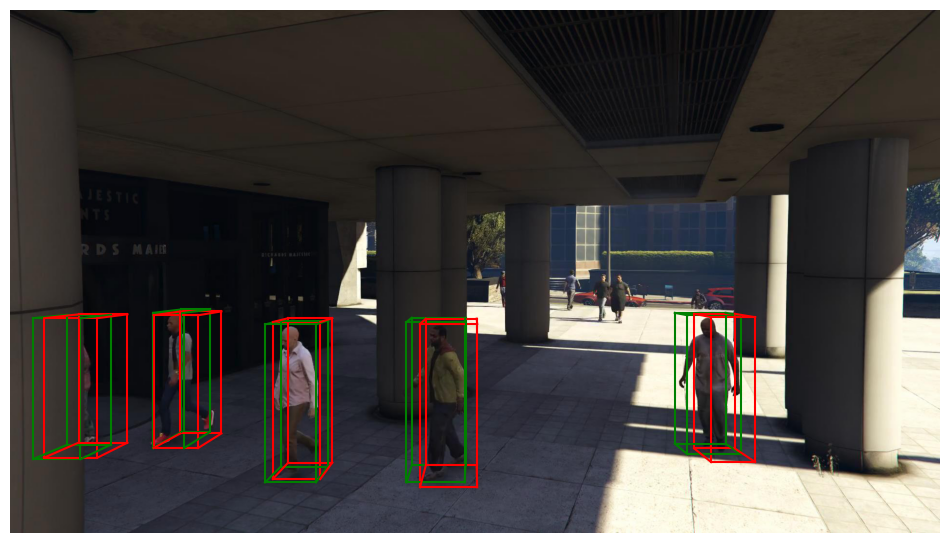}
\includegraphics[width=0.24\textwidth]{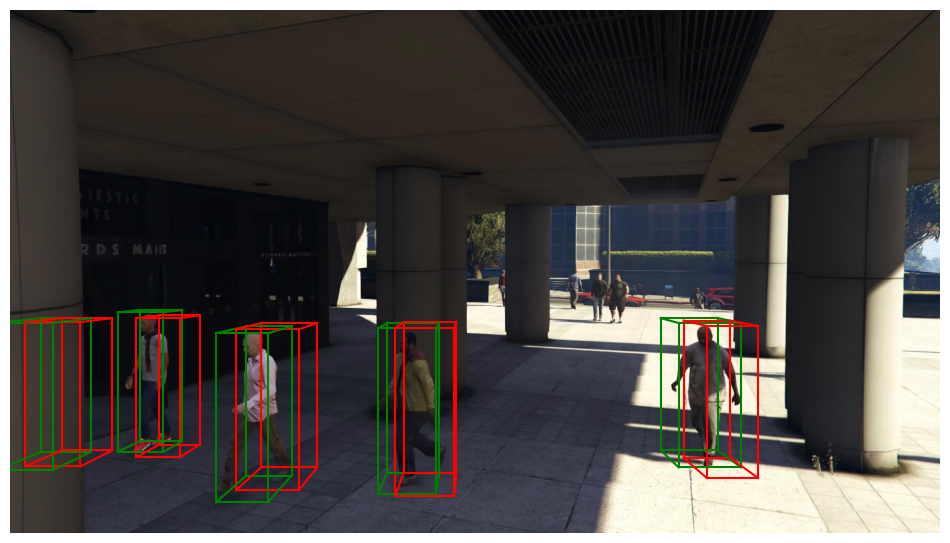}
\includegraphics[width=0.24\textwidth]{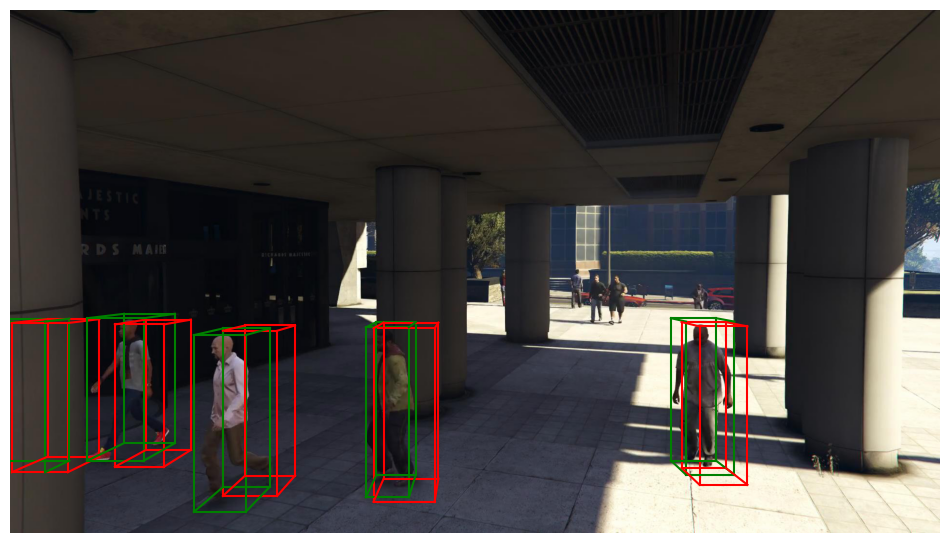}\\
\includegraphics[width=0.24\textwidth]{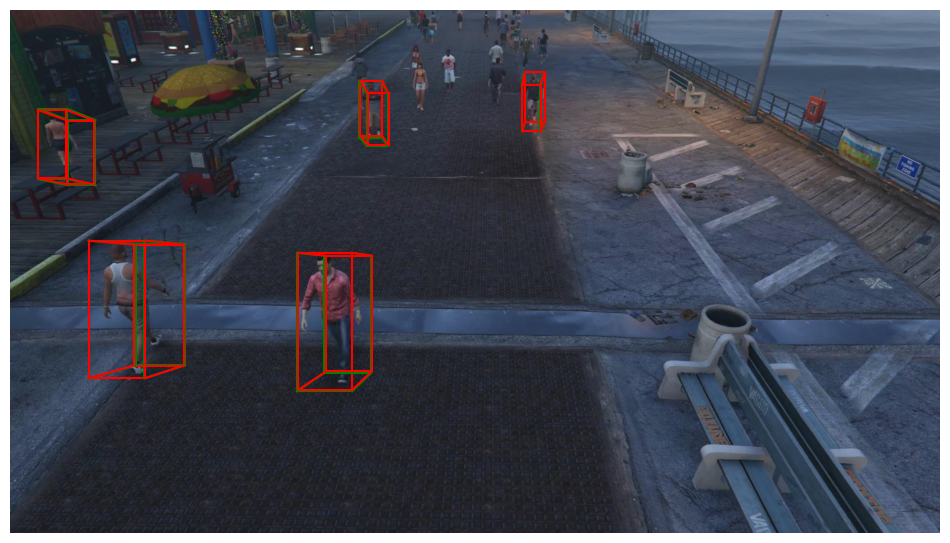}
\includegraphics[width=0.24\textwidth]{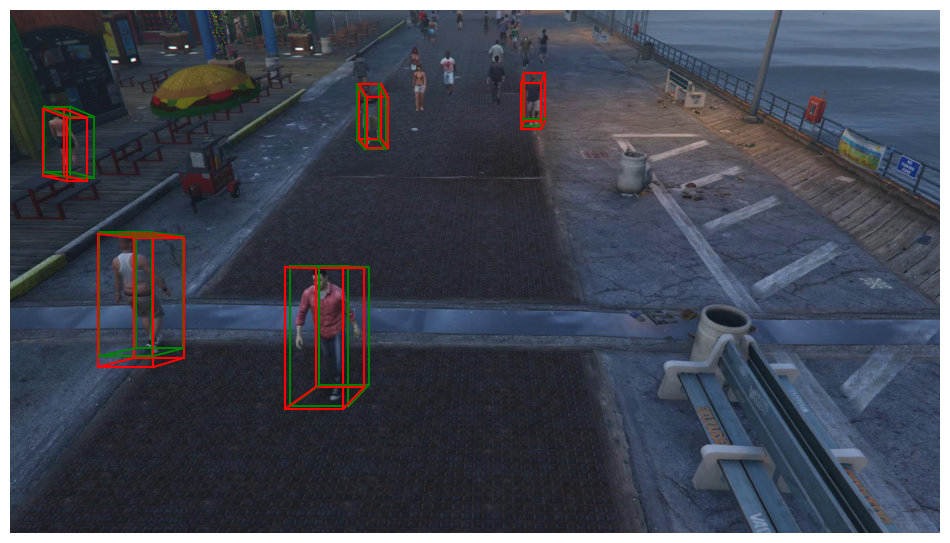}
\includegraphics[width=0.24\textwidth]{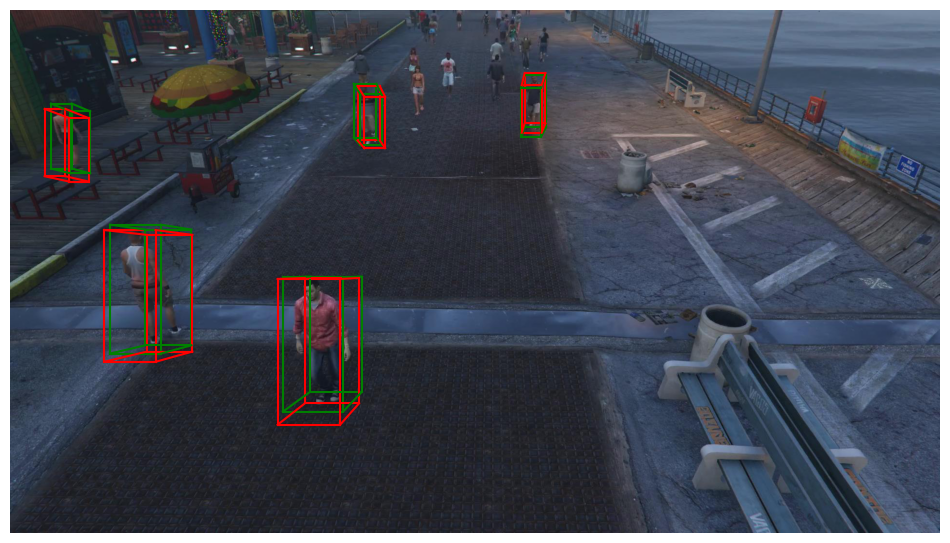}
\includegraphics[width=0.24\textwidth]{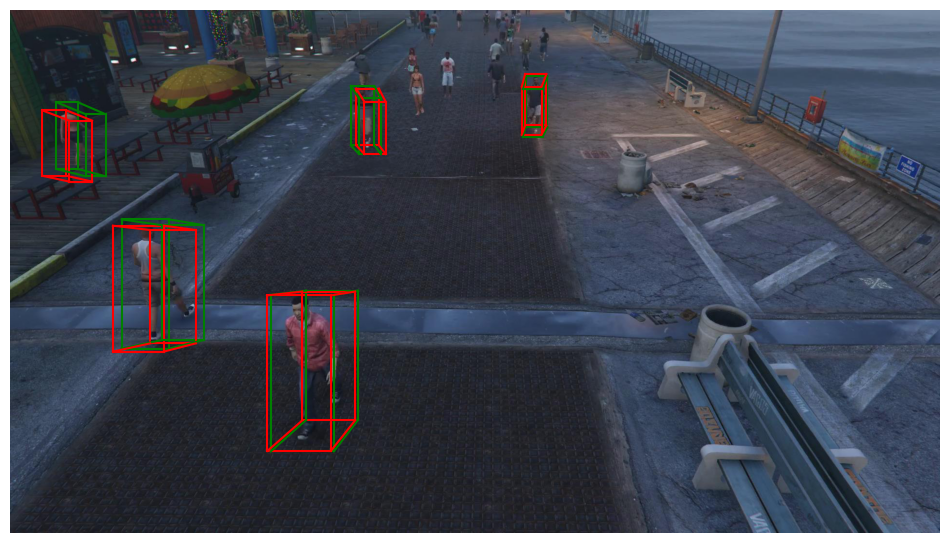}\\
\includegraphics[width=0.24\textwidth]{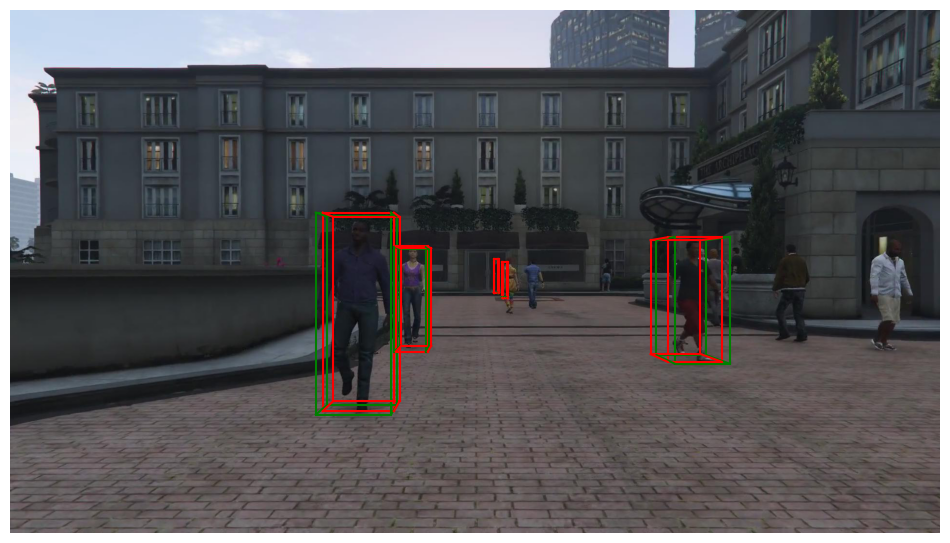}
\includegraphics[width=0.24\textwidth]{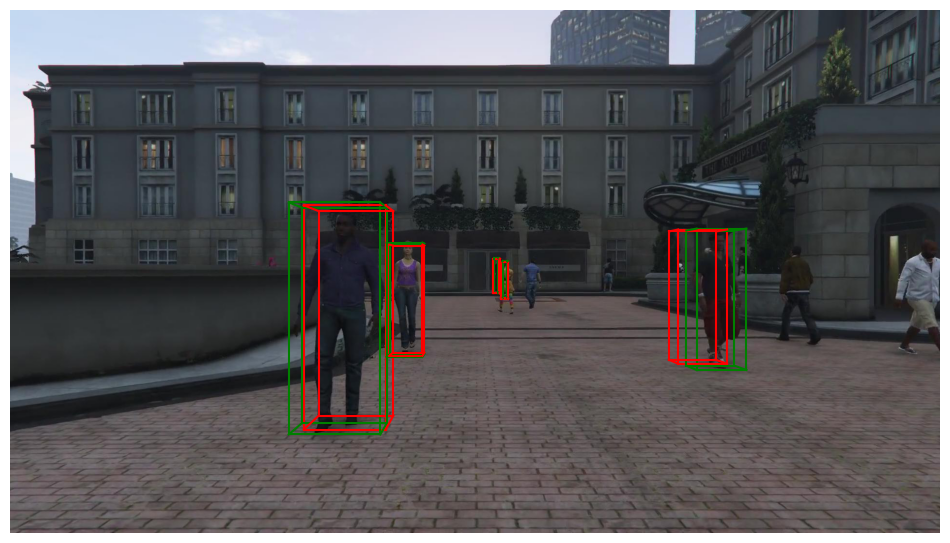}
\includegraphics[width=0.24\textwidth]{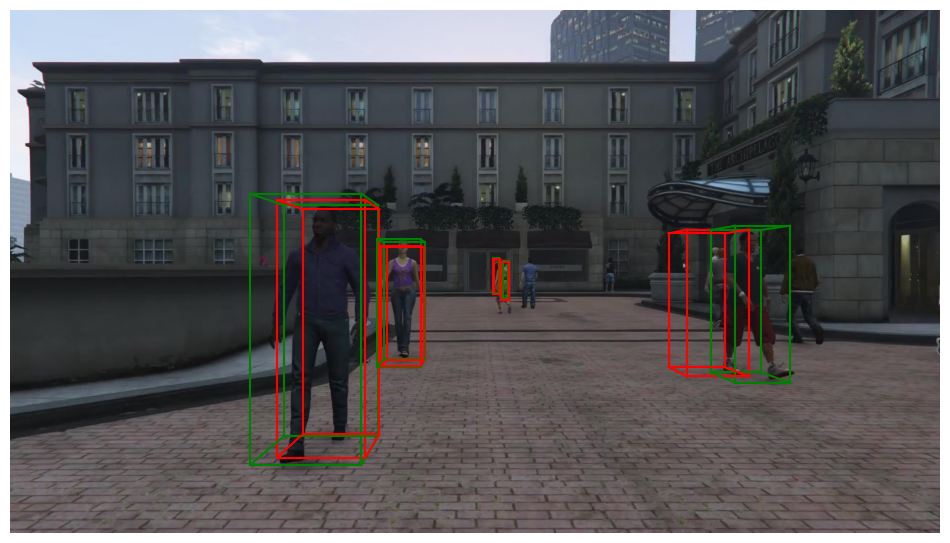}
\includegraphics[width=0.24\textwidth]{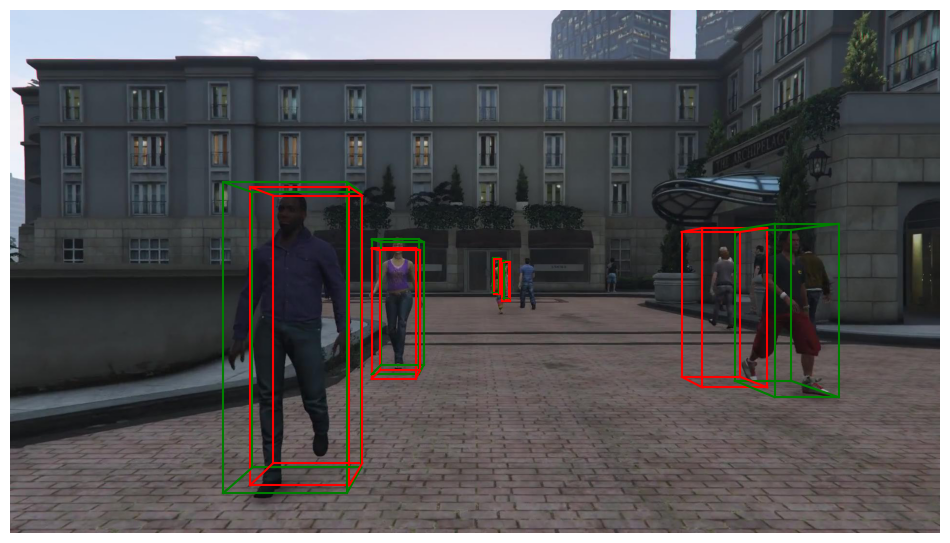}
\caption{Qualitative results on JTA dataset. From left to right, every 4 frames of predictions are depicted. The green is the ground truth and the red is the predicted bounding box.}
\label{fig:jta}
\end{figure}

Qualitatively, the predicted bounding boxes tend to be close to the ground truth bounding boxes, as shown in \Cref{fig:jta}. The network can predict accurately when a pedestrian is standing. However, it sometimes misinterprets the speed at which the pedestrian is moving, causing the predicted bounding boxes to either lead or lag the ground truth.

The performance of the model on the NuScenes dataset is shown in \Cref{tab:nuscenes}. AIOU and FIOU scores are lower compared to the JTA dataset, and it may be caused by fewer observations (only $4$ frames) compared to the JTA dataset. For example, the network seems to have more difficulties predicting the bounding boxes of pedestrians that are further away from the camera.

\begin{table}[!h] 
    \centering
    \caption{Quantitative results on NuScenes dataset. For ADE/FDE, the lower the better and for AIOU/FIOU, the higher the better.}
    \begin{tabular}{c| c c c c}
        \toprule
        & \multicolumn{4}{c}{prediction horizon = 2 sec} \\
        \midrule
        Model & ADE & FDE & AIOU & FIOU \\
        \midrule
        Zero-Vel & 4.3645 & 6.9420 & 0.0843 & 0.0652 \\
        P-LSTM & 1.0944 & 1.9851 & 0.2515 & 0.1100 \\
        PV-LSTM & \textbf{0.9945} & \textbf{1.8116} & \textbf{0.3059} & \textbf{0.1588} \\
        \bottomrule
    \end{tabular}
    \label{tab:nuscenes}
\end{table}

\begin{figure}[!h]
\centering
\includegraphics[width=0.24\textwidth]{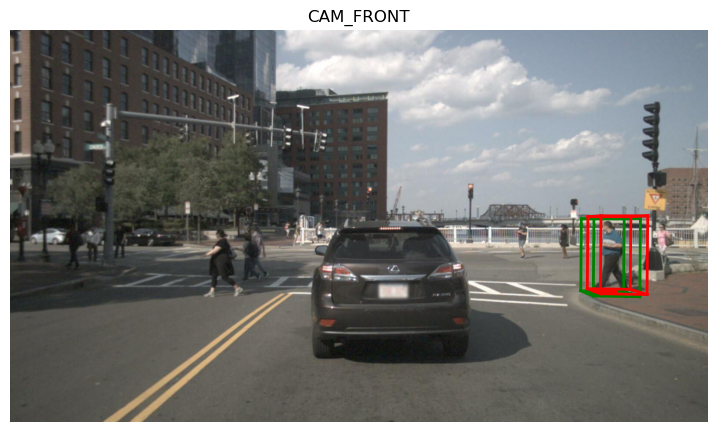}
\includegraphics[width=0.24\textwidth]{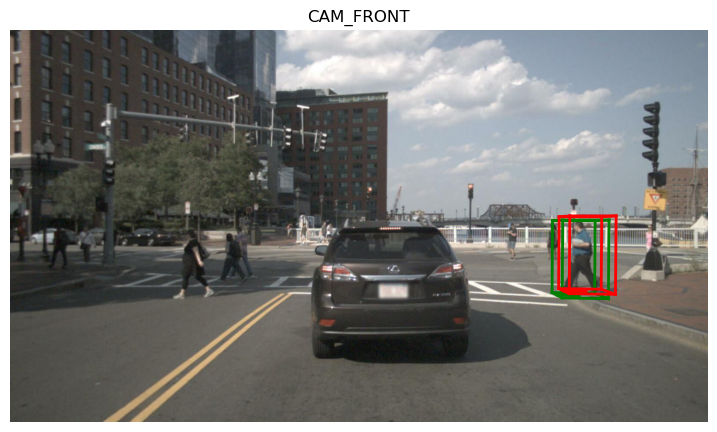}
\includegraphics[width=0.24\textwidth]{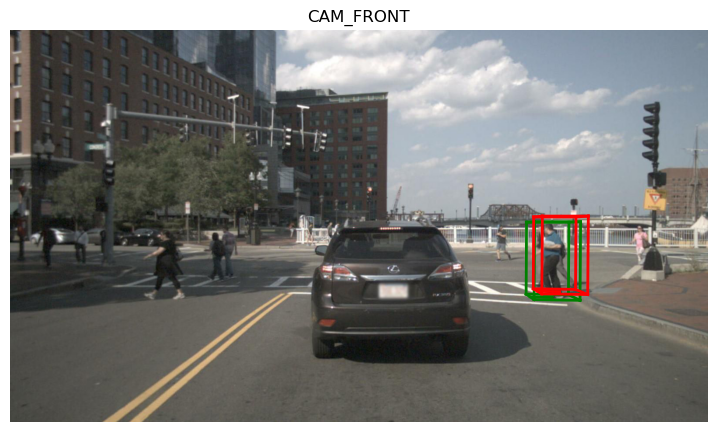}
\includegraphics[width=0.24\textwidth]{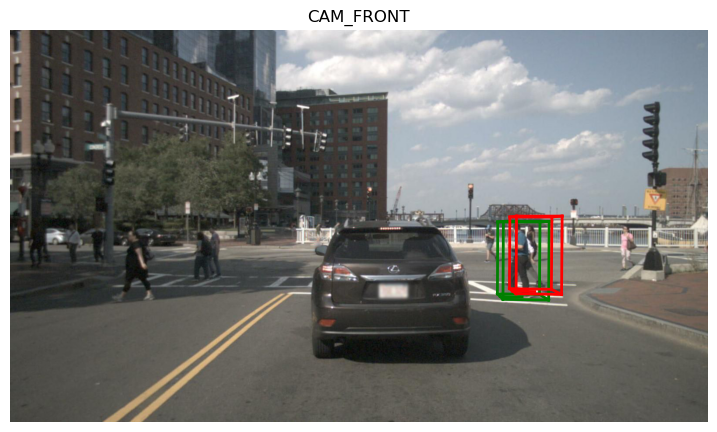}\\
\includegraphics[width=0.24\textwidth]{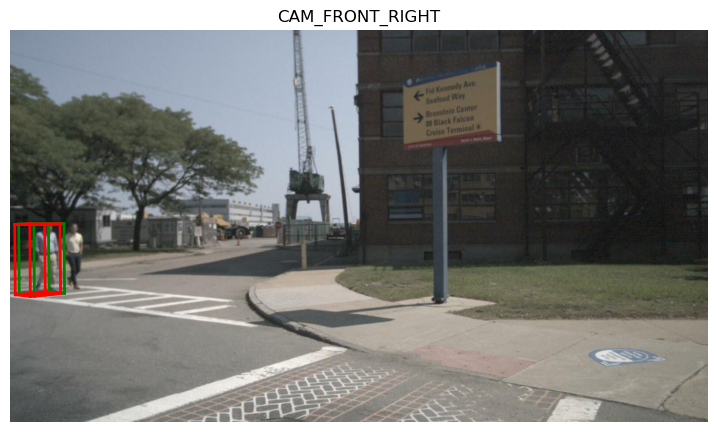}
\includegraphics[width=0.24\textwidth]{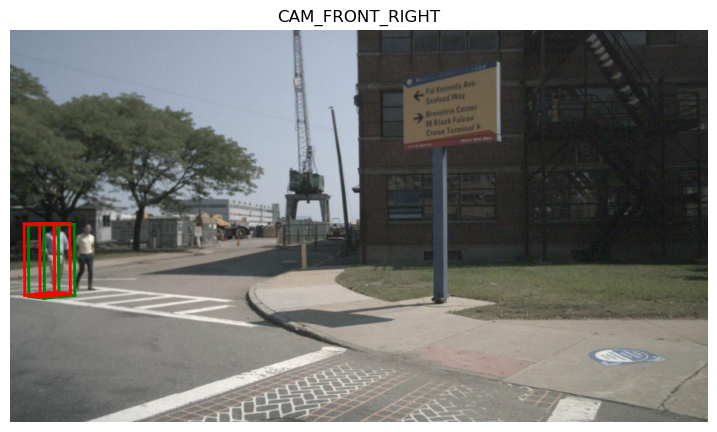}
\includegraphics[width=0.24\textwidth]{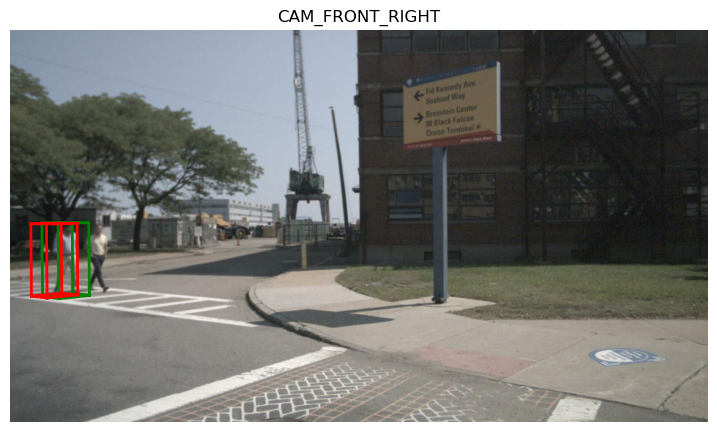}
\includegraphics[width=0.24\textwidth]{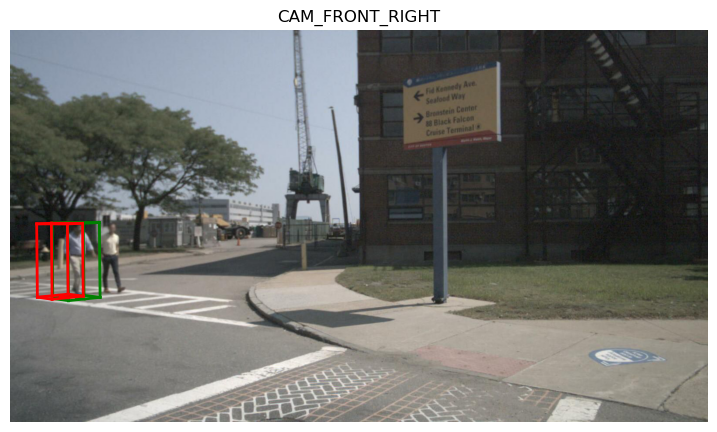}\\
\includegraphics[width=0.24\textwidth]{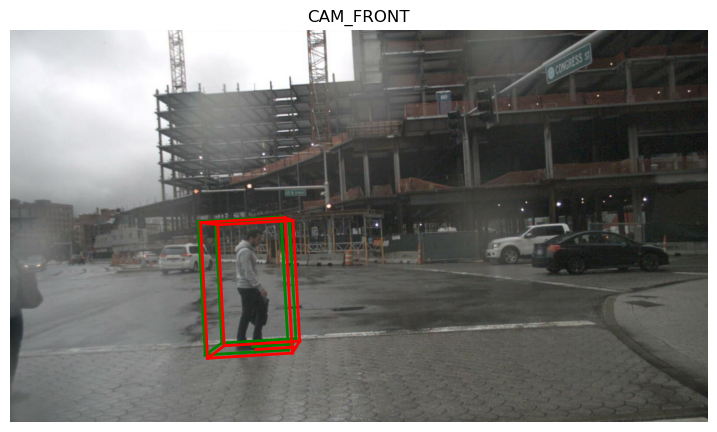}
\includegraphics[width=0.24\textwidth]{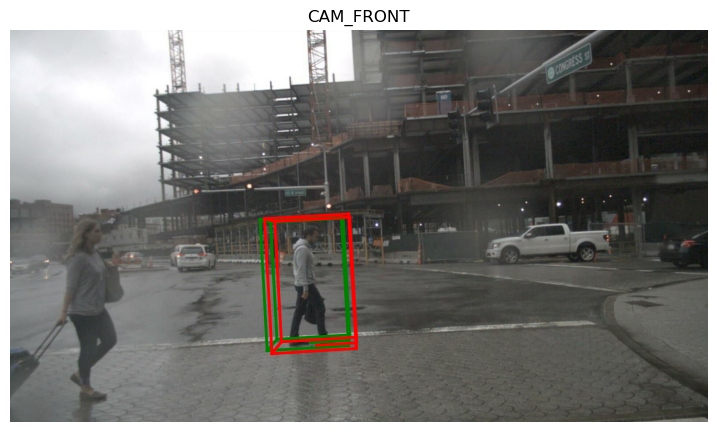}
\includegraphics[width=0.24\textwidth]{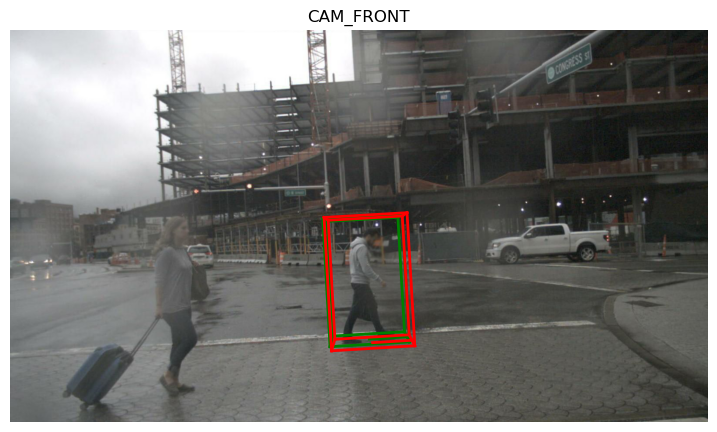}
\includegraphics[width=0.24\textwidth]{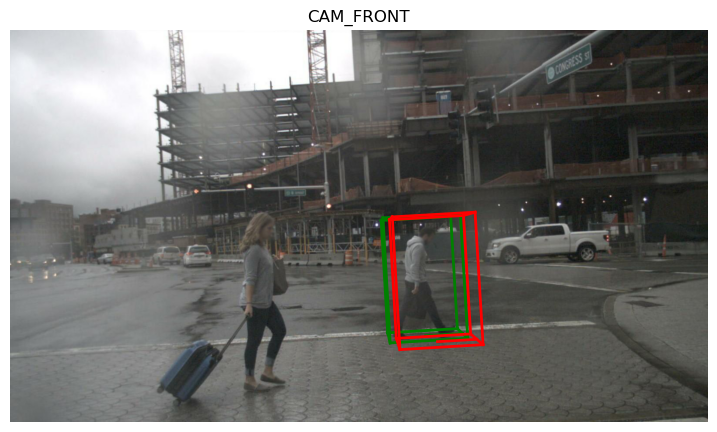}\\
\includegraphics[width=0.24\textwidth]{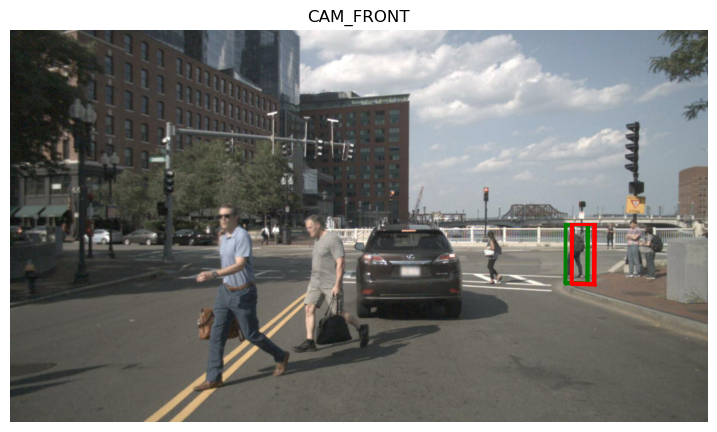}
\includegraphics[width=0.24\textwidth]{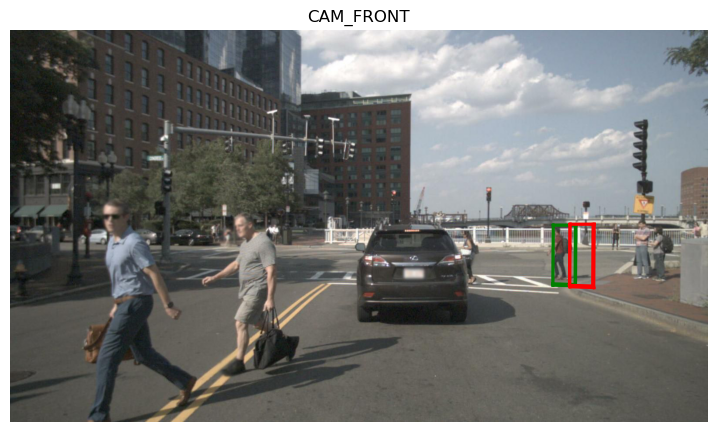}
\includegraphics[width=0.24\textwidth]{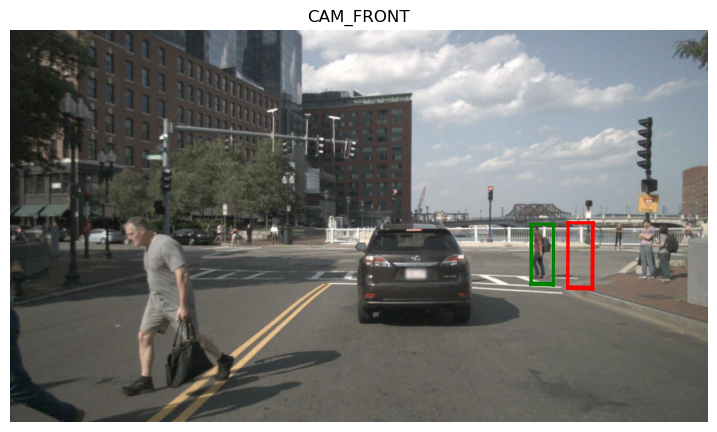}
\includegraphics[width=0.24\textwidth]{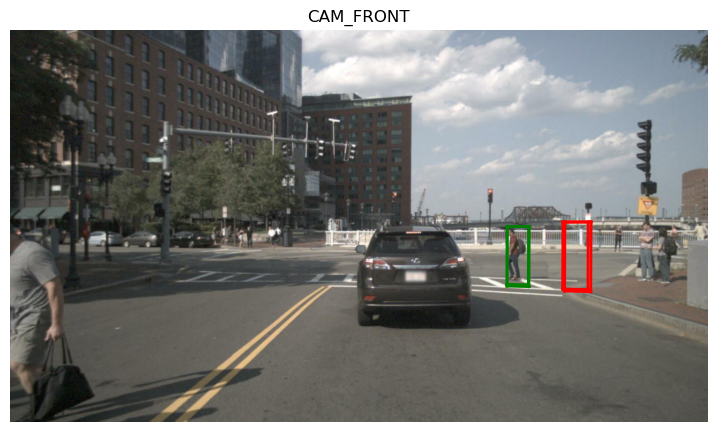}\\
\caption{Qualitative results on NuScenes dataset. From left to right, all predictions are depicted. The green is the ground truth and the red is the predicted bounding box.}
\label{fig:nuscenes}
\end{figure}

Qualitative results show that the network successfully tracks the location of pedestrians in many instances. The network usually accurately predicts the locations of people, as shown in rows 1-3 in \Cref{fig:nuscenes}. One unsuccessful example is shown in the last row of \Cref{fig:nuscenes} where the predicted 3D bounding boxes tend to drift far away from the ground truth.

On the NuScenes dataset, there are annotations for human actions. Specifically, attribute labels indicate whether a pedestrian is moving, standing, or sitting/lying down. To validate our claim that the learned representation is also useful for other tasks, we predict the 3D bounding box along with action (multi-task PV-LSTM) and compare it with the network when there is no 3D bounding box prediction decoder (single-task PV-LSTM).
We balanced the data for better training, and the network was penalized only by the final prediction.
The results are in \cref{tab:action}. The multi-task model has more than $2\%$ higher accuracy in action anticipation. 

\begin{table}[!h] 
    \centering
    \caption{Action anticipation accuracy on NuScenes dataset.}
    \begin{tabular}{c| c}
        \toprule
        Model & action prediction \\
        \midrule
        single-task PV-LSTM & 80.8\% \\
        multi-task PV-LSTM & \textbf{82.9\%} \\
        \bottomrule
    \end{tabular}
    \label{tab:action}
\end{table}

\section{Conclusion and Future Work}
Predicting pedestrian motion and actions in the wild continues to be an essential topic for the advancement and widespread adoption of autonomous vehicles. Most research in this area is focused on 2D predictions using trajectories, pose keypoints, or bounding boxes. However, data in 3D can provide more information on the pedestrians for autonomous vehicles. Here, we proposed an encoder-decoder LSTM network for 3D human bounding box predictions. It can provide a wealth of information on pedestrian location and size in the near future in 3D so that autonomous vehicles can make better and safer decisions. We showed that the learned representation is also helpful for action anticipation.

With the emergence of more 3D datasets for various autonomous driving tasks and more annotations, the usefulness of the learned representation on other tasks such as pedestrian crossing intention detection can be investigated. This framework can also be extended to other agents such as vehicles on the road and other datasets in future work. In addition, modeling the camera movement explicitly and excluding it from human motions are left for future studies.

\section*{Acknowledgements}
This project has received funding from the European Union's Horizon 2020 research and innovation program under the Marie Sklodowska-Curie grant agreement No 754354.

\bibliographystyle{apacite}
\bibliography{sample}

\end{document}